# Biogeography based Satellite Image Classification

V.K.Panchal
Director, DTRL
DRDO,
Delhi, INDIA

Parminder Singh
Assistant Professor
G.N.E College,
Ludhiana, Punjab, INDIA

Navdeep Kaur
M.Tech Student,
G.N.E College,
Ludhiana, Punjab, INDIA

Harish Kundra
Assistant Professor
R.I.E.I.T,
Railmajra, Punjab, INDIA

*Abstract-* **Biogeography is the study of the geographical distribution of biological organisms. The mindset of the engineer is that we can learn from nature. Biogeography Based Optimization is a burgeoning nature inspired technique to find the optimal solution of the problem. Satellite image classification is an important task because it is the only way we can know about the land cover map of inaccessible areas. Though satellite images have been classified in past by using various techniques, the researchers are always finding alternative strategies for satellite image classification so that they may be prepared to select the most appropriate technique for the feature extraction task in hand. This paper is focused on classification of the satellite image of a particular land cover using the theory of Biogeography based Optimization. The original BBO algorithm does not have the inbuilt property of clustering which is required during image classification. Hence modifications have been proposed to the original algorithm and the modified algorithm is used to classify the satellite image of a given region. The results indicate that highly accurate land cover features can be extracted effectively when the proposed algorithm is used.**

*Keywords – Biogeography, Biogeography Based Optimization, Satellite Image Classification, Rough Set Theory*

## I. INTRODUCTION

Biogeography is the study of geographic distribution of biological organisms. The science of Biogeography can be traced to the work of century nationalists such as Alfred Wallace [9] and Charles Darwin [2]. Biogeography based Optimization (BBO) is an application of biogeography to optimization problems. It is modeled after the immigration and emigration of species between the islands.

*A. Biogeography based Optimization*

Biogeography Based Optimization (BBO) is a population based evolutionary algorithm (EA) motivated by the migration mechanisms of ecosystems. It is based on the mathematics of biogeography. In BBO, problem solutions are represented as islands, and the sharing of features between solutions is represented as emigration and immigration. BBO was first presented in December 2008 by D. Simon [1]. It is an example that a natural process can be modeled to solve general optimization problems.

One characteristic of BBO is that the original population is not discarded after each generation. It is rather modified by migration. Also for each generation, BBO uses the fitness of each solution to determine its emigration and immigration rate. In a way, we can say that BBO is an application of biogeography to EAs. Other distinguishing feature of BBO is that when updating the population, BBO considers the fitness of the immigrating and emigrating islands via immigration and emigration curves.

In BBO, each individual solution is considered as a habitat with a habitat suitability index (HSI), which is similar to the fitness of EAs, to measure the suitability of individual. Also, an SIV (suitability index variable) which characterizes the habitability of an island is used. A good solution is analogous to an island with a high HSI, and a poor solution indicates an island with a low HSI. High HSI solutions tend to share their features with low HSI solutions. Low HSI solutions accept a lot of new features from high HSI solutions. In BBO, each individual has its own immigration rate $\lambda$ and emigration rate $\mu$. A good solution has higher emigration rate $\mu$ and lower immigration rate $\lambda$.

*B. Rough Set Theory*

In the rough set approach, it is assumed that any vague concept is replaced by a pair of precise concepts called the lower and the upper approximation of the vague concept. The lower approximation consists of all objects that surely belong to the concept and the upper approximation contains all objects that possibly belong to the concept. Obviously, the difference between the upper and the lower approximation constitutes the boundary region of the vague concept [8]. Discretization process of rough set theory can be described as one that returns a partition of the value sets of conditional attributes into intervals. The partition is done in such a way that if the name of the interval containing an arbitrary object is substituted for any object instead of its original value in decision table, a consistent decision is also obtained. In this way the size of the value attribute sets in a decision system is reduced.

*C. Remote Sensing Image Classification*

In remote sensing, satellite based sensors are burgeoning as a major facilitator of geo-spatial information providing different manifestations of the terrain. The satellite image is one of the main sources for capturing the geo-spatial information. Remote sensing with multi spectral satellite imagery is based on the concept that different features/objects constituting the land cover reflect electro-magnetic radiations over a range of wavelengths in its own characteristics way according to its chemical composition and physical state. A multi-spectral remote sensing system operates in a limited number of bands and measures radiations in series of discrete spectral bands. The spectral response is represented by the discrete digital number (DN). Spectral signatures of an object may be used for identification much like a fingerprint [7]

There are two main type of classifying techniques: *Supervised and Unsupervised classification* [5]. When spectral classes, based on numerical information, are grouped first and are then matched by the analyst to information classes, then it is termed as unsupervised classification. Clustering algorithms are used to determine the statistical structures in the data for example K-Means approach. In supervised classification, the homogeneous samples of the different surface cover types of interest are used. To recognize spectrally similar areas, numerical information in all spectral bands for the pixels comprising these areas is used. For each pixel in the image a comparison is made with these signatures and defined in the class it most closely "resembles".

## II. PROPOSED WORK

In the proposed work, Biogeography Based Optimization Algorithm has been used to classify a satellite image. Biogeography based Optimization algorithm is a basically used to find the optimal solution of a problem. But satellite image classification is a clustering problem that requires each class to be extracted as a cluster. The original BBO algorithm does not have the inbuilt property of clustering. But here to extract features from the image, we have tried to make the clusters of different classes present in the image and proposed a modified Biogeography based algorithm classify a satellite image.

For the purpose of feature extraction, 7-band satellite image of Alwar city (Rajasthan) of size 472 X 546 has been used. The BBO parameters of the proposed algorithm are defined as follows:

Definition 1: In our study, each of the multispectral band of image represents one Suitability Index Variable (SIV) of the habitat. Further, since image in each band is a gray image, SIV∈ C is an integer and C ⊂ [0,255].

Definition 2: A habitat H ∈ $SIV^m$ where m is 7 because we have 7 – band data.

Definition 3: Initially there exists a universal habitat that contains all the species to be migrated. Also we have as many other habitats as the number of classes to be found from the image. In our image, we want to classify 5 features –water, vegetation, urban, rocky and barren from the image. Each of these features is represented by a habitat. So the ecosystem $H^6$ is a group of 6 habitats (one universal habitat and five feature habitat).

Definition 4: Since we are using an image of almost 2.5 lakh pixels, considering each pixel as a species would make the proposed algorithm extremely slow. So we has used rough set theory to obtain the random clusters of pixels(by using discretization and partitioning concept of rough set theory) and each of the resulting cluster will be considered as mixed species that migrate from one habitat to another. These species can also be termed as 'elementary classes' of a habitat.

Definition 5: we have used standard deviation of pixels as Habitat Suitability Index to help in image classification.

Definition 6: The BBO algorithm [1] proposed the migration of SIV values from a high HSI habitat to a low HSI habitat. The shared features (SIV) remain in the high HSI solutions, while at the same time appearing as new features in the low HSI solutions.

In our proposed algorithm, rather than moving SIV, we are moving species altogether from a universal habitat to feature habitat. The species does not remain shared: it is removed from the universal habitat and migrated to feature habitat.

Definition 7: Maximum Immigration rate and Maximum Emigration Rate are same and equal to number of species in the habitat. The algorithm follows a linear curve (E=I). Number of species and thus the Maximum Immigration Rate and Maximum Emigration rate can vary in each iteration. Maximum species count ($S_{max}$) and the maximum migration rates are relative quantities. That is, if they all change by the same percentage, then the behavior of BBO will not change [1].

Definition 8: Since mutation is not an essential feature of BBO [6], it is not required in the proposed algorithm. Elitism, too, is an optional parameter [3]; it has not been used in the proposed algorithm.

So the assumptions made in this work can be summarized as follows:

(a). It has been assumed that species migrate from one habitat to another habitat as a mixed population.

(b) Also for the feature extraction purpose, in each generation all the habitats have been considered exactly once.

(c) In next generation only the unclassified pixels have been used. However the total population is still present.

(d) NIR and MIR band of image have been used for discretization of data. However, the bands used for discretization effects the final results

*A. Proposed Algorithm*

The following Biogeography based algorithm has been proposed to classify the satellite image:

```
Algorithm BiogeographyBasedSatelliteImageClassification

1. Get the Multispectral satellite image
2. Cluster the image randomly (using rough set theory) and
   consider each cluster as a species of universal habitat.
3. Consider other habitats one of each Land cover feature-
   Water, Urban, Rocky, Barren and Vegetation-having
   training pixels (produced by experts) as their members.
4. Define HSI, S_max, immigration rate (λ) and emigration rate
   (μ).
5. Calculate HSI of each feature habitat.
6. (i) Select a species from the Universal Habitat and migrate
   it to one of the feature habitat.
   (ii) Recalculate the HSI of feature habitat after the migration
   of the species to it.
7. If the recalculated HSI is within the threshold value(taken
   in the range -1 to +1 in our algorithm), then:
       (i) Absorb the species in the feature habitat.
       (ii) Go to step 8.
   Else if any unconsidered feature habitat is left then:
       (i) Migrate the species to that feature habitat.
       (ii) Go to step 7.
   Else
       (i) Use the rough set theory to discretize species (as it
       contained mixed pixels) and make random clusters
       which is considered as separate species.
       (ii) Add these new species to universal habitat.
8. If no species is left in the universal habitat then:
       (i) Stop the process.
   Else:
       (i) Go to step 6.
   End
```

Figure 1: Algorithm for Biogeography Based Satellite Image Classification

Initially random clusters of pixels are obtained using rough set theory. Each cluster is put in the universal habitat, considering each cluster as a species (or elementary class) in universal habitat. The feature habitats initially contain training pixels and HSI is calculated on these training pixels. Then each of the species is migrated to feature habitat one by one and HSI of the habitat is recalculated after migration. If HSI of habitat changes beyond a certain threshold limit, it means that the habitat is not suitable for that species and it should be migrated to some other feature habitat. If no feature habitat is found suitable for the species, the species is put back in universal habitat. This also means that that species is actually a mixture of more than one species and those needs to be separated from each other using discretization in rough set theory.

### III. RESULTS AND DISCUSSIONS

Our objective is to use the proposed Biogeography based algorithm as an efficient Land cover classifier for satellite image. We have taken a multi-spectral, multi resolution and multi-sensor image of Alwar area in Rajasthan. The satellite image for 7different bands is taken (Fig 2). These bands are Red, Green, Near Infra Red(NIR), Middle Infra Red (MIR), Radarsat-1 (RS1), Radarsat-2(RS2), and Digital Elevation Model) DEM.

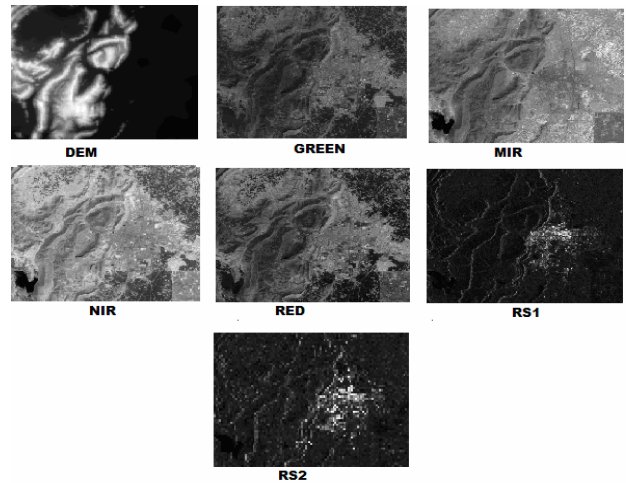

Figure 2: Seven Band Image of Alwar (Rajasthan)

Initially each feature habitats contain training pixels of its class as shown in figure 3. HSI of feature habitat is calculated on these pixels.

Figure 3: Training pixels originally present in each habitat

Then each species in the universal habitat are migrated to feature habitat and HSI is recalculated after migration to find the suitability of species in that feature habitat as shown in the figure 4. A given species is absorbed (and hence it belongs) to that feature habitat where it makes minimum variation in HSI after migration. For e.g. in the figure 4, 'pure rocky' represents HSI calculated with training pixels in rocky habitat and 'with rocky' represents HSI recalculation after migrating the species to this habitat. Since the difference between the two HSI values is minimal, the species will be migrated to 'Rocky' habitat.

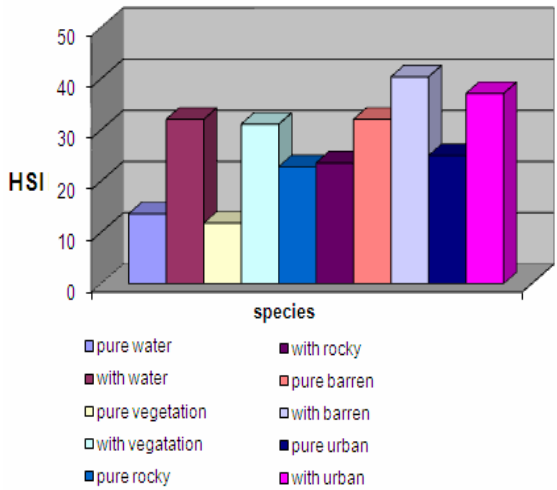

Figure 4: HSI matching of a given species in each Feature habitat

After applying the proposed algorithm to the 7-band of Alwar Image, the classified image is obtained in figure 5. The yellow, black, blue, green, red color represents rocky, barren, water, vegetation, urban region respectively.

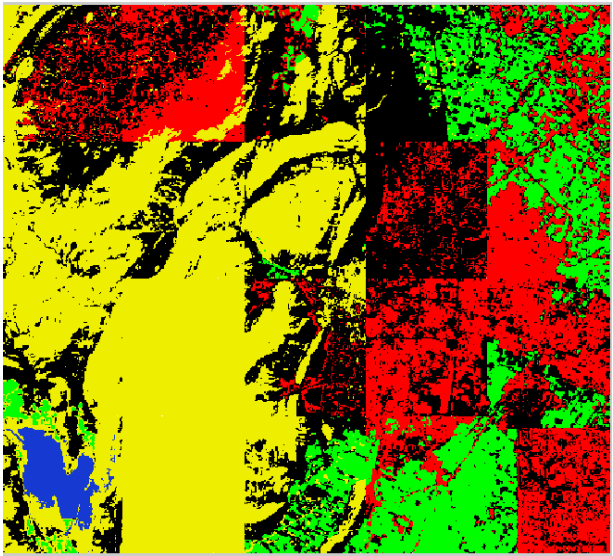

Figure 5: Biogeography based classification of satellite image of Alwar (with K-coefficient=0.6715)

As the threshold limit of HSI matching is lowered, the species do not get absorbed in the feature habitat and return to universal habitat. Those species are further discretized and classified in next iterations (generation). As the number of iterations is increased, we get more refined results as shown in figure 6. By taking threshold value from -1 to +1, no water body (represented in blue color) is extracted in first iteration (fig.6 (a)) and water pixels remain unclassified (represented in white color). With the increasing iteration (fig. 6(b), 6(c), 6(d), 6(e)), water body is clearly extracted at the end of fifth iteration.

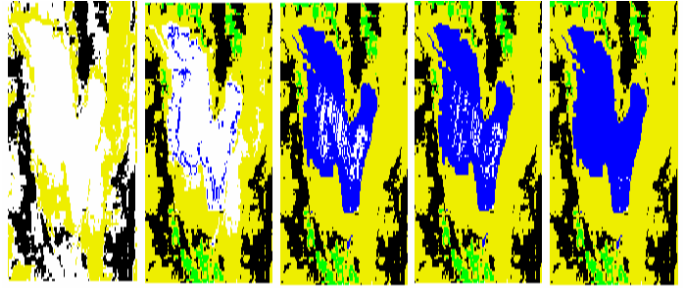

(a)　　　　(b)　　　　(c)　　　　(d)　　　　(e)

(a) Iteration 1   (b) Iteration 2   (c) Iteration 3   (d) Iteration 4   (e) Iteration 5

Figure 6: Water Body extraction (in blue color) from the Alwar image with each iteration. White color portion represent unclassified image

A. Accuracy Assessment

A classification is not complete until its accuracy is assessed [4] Classification accuracy of our proposed algorithm is expressed using classification *error matrix*. Error matrices compare, on a category-by category basis, the relationship between known reference data (ground truth) and the corresponding results of an automated classification. We took 150 vegetation pixels, 190 Urban pixels, 200 Rocky pixels, 70 water pixels, 170 barren pixels from the training set and the error matrix obtained is shown figure 7. The error matrix's interpretation along column suggests how many pixels are classified correctly by algorithm. For e.g. in the first column, out of total 150 vegetation pixels, 127 pixels were correctly classified into vegetation by the proposed algorithm, 6 were misclassified as rocky and 17 were misclassified as Barren .

|  | Vegetation | Urban | Rocky | Water | Barren | Total |
|---|---|---|---|---|---|---|
| Vegetation | 127 | 9 | 0 | 0 | 2 | 138 |
| Urban | 0 | 88 | 1 | 0 | 32 | 121 |
| Rocky | 6 | 2 | 176 | 1 | 17 | 202 |
| Water | 0 | 0 | 3 | 69 | 0 | 72 |
| Barren | 17 | 91 | 20 | 0 | 119 | 247 |
| Total | 150 | 190 | 200 | 70 | 170 | 780 |

Figure 7: Error Matrix of Biogeography based Satellite Image Classification of Alwar Region

The Kappa coefficient of the Alwar image can be calculated by applying following formula to the Error Matrix:

.

$$\hat{k} = \frac{N\sum_{i=1}^{r} x_{ii} - \sum_{i=1}^{r}(x_{i+} \cdot x_{+i})}{N^2 - \sum_{i=1}^{r}(x_{i+} \cdot x_{+i})} \quad (1)$$

**r** = number of rows in the error matrix (**r** =5 in our case)

***x*** ii = the number of observations in row i and column i (on the major diagonal)

***x*** i+ = total of observations in row i (shown as marginal total to right of the matrix)

***x*** +i = total of observations in column i (shown as marginal total at bottom of the matrix)

**N** = total number of observations included in matrix.(N=780 in our case)

The Kappa (K) coefficient of the Alwar image is 0.6715 which indicates that an observed classification is 67 .15 percent better than one resulting from chance.

## CONCLUSION AND FUTURE WORK

In this paper we have presented a Biogeography based algorithm as an efficient Land cover classifier for satellite image. The results presented are preliminary and there is a lot of scope for improvement to develop this algorithm as efficient classifier. From the results it can be seen that the refined results can be obtained by lowering the threshold value of HSI matching and hence increasing the number of iterations of the algorithm. It is also observed that the partitions generated by rough sets, do affect the final results.

The future scope of the research includes proposing certain modification to the algorithm so that the Kappa coefficient can be improved further. Also, an unsupervised version of the Biogeography based satellite image classification may be explored by including the context information of clusters used in generation of species granules and then evaluating the HSI.

## AUTHORS PROFILE

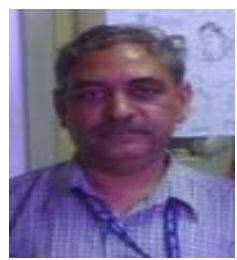
V.K. Panchal is Director at Defence Terrain Research Lab, New Delhi. Associate Member of IEEE (Computer Society) and Life Member of Indian Society of Remote Sensing. Chaired sessions & delivered invited talks at many national & international conferences. Research interest are in synthesis of terrain understanding model based on incomplete information set using bio-inspired intelligence and remote sensing.

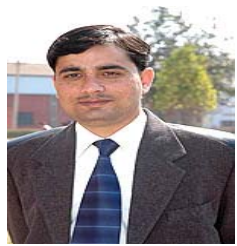
Harish Kundra is an Assistant Professor and Head of Department in the Computer Science and IT Department of Rayat Institute of Engineering & IT, Railmajra, Punjab, India. He graduated from Baba Banda Singh Bahadur Engineering College , Fatehgarhsahib in Computer Science & Engineering in 2000 and received Master's Degree in Computer Science and Engineering from Vinayka Mission Research Foundation ( Deemed University), Tamil naidu in 2007 . He has presented 15 papers in International/National Journals and 50 papers in International/National Conferences. He is a lifetime member of Indian Society of Technical Education. He guided more than 100 projects at graduate and master's level and also 4 dissertations at Master's level

.

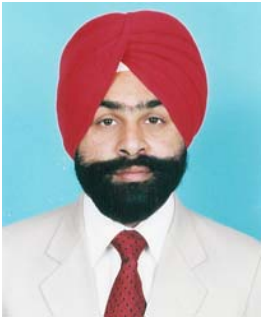
Parminder Singh is Assistant Professor in department of Computer Science and Engineering at Guru Nanak Dev Engineering College, Ludhiana(India). He is having about 10 years academics experience. He has guided about 10 post graduate students for their dissertation work. Presently he is working for the development of a Text-to-Speech synthesis system for Punjabi language.

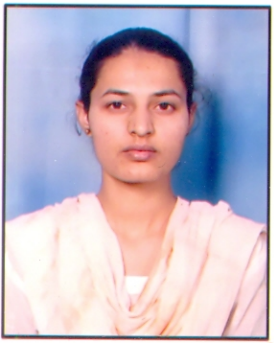
Navdeep Kaur has done B-Tech (Hons.) in Computer Science & Engineering & scored 81% marks from Punjab Technical University, Jalandhar (India) in 2005 and M-Tech in Computer Science & Engineering from Guru Nanak Dev Engineering College, Ludhiana of India in 2009. She is currently working as a lecturer in computer science and IT department of Rayat institute of Engineering & Information Technology of India.